Original Article

# CURVELET-BASED FREQUENCY-AWARE FEATURE ENHANCEMENT FOR DEEPFAKE DETECTION

**Salar Adel Sabri[1,*], Ramadhan J. Mstafa[1, 2]**

[1] Department of Computer Science, College of Science, University of Zakho, Zakho, Kurdistan Region, Iraq.
[2] PRIME Lab, Scientific Research Center, University of Zakho, Zakho, Kurdistan Region, Iraq.

*Corresponding Author, E-mail:* salaradelsabry@gmail.com (Tel: +964-770-7176353)

**ABSTRACT**

Received: 03, Aug, 2025
Accepted: 19, Sep, 2025
Published: 13, Apr, 2026

The proliferation of sophisticated generative models has significantly advanced the realism of synthetic facial content, known as deepfakes, raising serious concerns about digital trust. Although modern deep learning-based detectors perform well, many rely on spatial-domain features that degrade under compression. This limitation has prompted a shift toward integrating frequency-domain representations with deep learning to improve robustness. Prior research has explored frequency transforms such as Discrete Cosine Transform (DCT), Fast Fourier Transform (FFT), and Wavelet Transform, among others. However, to the best of our knowledge, the Curvelet Transform, despite its superior directional and multiscale properties, remains entirely unexplored in the context of deepfake detection. In this work, we introduce a novel Curvelet-based detection approach that enhances feature quality through wedge-level attention and scale-aware spatial masking, both trained to selectively emphasize discriminative frequency components. The refined frequency cues are reconstructed and passed to a modified pretrained Xception network for classification. Evaluated on two compression qualities in the challenging FaceForensics++ dataset, our method achieves 98.48% accuracy and 99.96% AUC on FF++ low compression, while maintaining strong performance under high compression, demonstrating the efficacy and interpretability of Curvelet-informed forgery detection.

## 1. INTRODUCTION

The rapid advancement of artificial intelligence, particularly in the area of generative modelling, has led to the emergence of highly convincing facial forgeries, commonly known as deepfakes. These synthetic images and videos are designed to imitate the appearance, expressions, and behaviors of real individuals with such realism that even human observers often struggle to distinguish them from authentic content (Goodfellow *et al.*, 2014; Khalid *et al.*, 2023; Mustak *et al.*, 2023).

The increasing accessibility of such tools raises urgent concerns regarding identity theft, misinformation, and digital trust (Khalid *et al.*, 2023; Mustak *et al.*, 2023). In response, numerous detection approaches have emerged to address these threats. Early efforts focused on handcrafted features, leveraging cues such as inconsistent blinking patterns or unnatural facial geometry to detect tampered content (Li & Lyu, 2019; Matern *et al.*, 2019). More recent developments have shifted toward deep learning-based approaches, where convolutional neural networks (CNNs) are trained to automatically learn discriminative spatial representations from manipulated facial regions (Abbas & Taeihagh, 2024; Cozzolino *et al.*, 2017). While these methods have demonstrated promising results in controlled settings, their performance often deteriorates when visual quality is compromised or when tested on unseen manipulation types.

To overcome these limitations, researchers have increasingly turned to the frequency domain. Spatial cues are especially vulnerable to compression and resolution loss, making RGB-based detectors less reliable in real-world scenarios (Gao *et al.*, 2024). In contrast, frequency-domain representations can reveal subtle, yet consistent artifacts introduced during generative synthesis, particularly through upsampling and blending operations (Frank *et al.*, 2020a; Huang *et al.*, 2020). Transforms such as the Fourier and Cosine transforms have been widely adopted to extract these signals and guide detection networks in identifying forgeries even when spatial artifacts are suppressed (Gao *et al.*, 2024; Tan *et al.*, 2024; Wolter *et al.*, 2022).

However, most of these frequency-based approaches rely on transforms that lack directional sensitivity and multiscale geometric localization, both of which are fundamental to accurately capturing facial contours and edge structures that are often disrupted in manipulated content. The Curvelet Transform, in contrast, is explicitly designed to represent image edges along curves and across multiple orientations and scales with high sparsity and fidelity (Emmanuel & David, 2000; Emmanuel & Guo, 2002).

In this work, we aim to explore and evaluate the use of Curvelet Transform in combination with deep learning techniques for deepfake detection. We investigate whether Curvelet-transformed facial signals encode meaningful differences between real and forged content, and how such signals can be integrated within a CNN-based network for detection purposes.

The main contributions of this research are summarized as follows:

- We explore Curvelet Transform for deepfake detection for the first time, leveraging its directional and multiscale properties to better capture facial geometry and edge artifacts than traditional frequency transforms.
- We propose **Curvelet-FAFE**, a novel method that separately processes curvelet magnitude and phase, incorporating **WedgeSE**, a spatially-aware, wedge-level attention module with binary gating, to dynamically select informative frequency components.
- We group curvelet responses into interpretable, scale-based bands modulated by learnable spatial masks, enabling structural inconsistency detection across resolutions.
- We introduce a progressively intensified **L1 regularization** on gate activations, dynamically adapting to classification loss and applied channel-wise to ensure balanced, compact feature selection across RGB.
- Extensive experiments on deepfake datasets show that curvelet-based features effectively capture forgery cues. Using a pretrained Xception backbone, our model achieves strong classification performance, with wedge activation patterns confirming its interpretability.

## Related work:

Deepfake detection methods can be broadly classified into spatial-domain and frequency-domain approaches. Below, we review representative works in each category, highlighting their strengths, limitations, and relevance to our Curvelet-based investigation.

## Spatial-Domain Forgery Detection:

Early deepfake detectors focused on hand-crafted cues extracted from RGB images or videos. For instance, some methods analyse colour inconsistencies and shape artifacts introduced during synthesis (José De Carvalho *et al.*, 2013; Carvalho *et al.*, 2016), while others exploit pixel-level error metrics, such as Error Level Analysis, to reveal tampering. Hand-crafted feature pipelines combined with shallow classifiers (e.g., SVMs) showed initial promise (Bayar & Stamm, 2016) but lacked scalability as forgery techniques grew more sophisticated.

The advent of deep learning shifted emphasis toward CNN-based detectors capable of learning hierarchical spatial features directly from data. MesoInception-4 (Afchar *et al.*, 2018), inspired by the Inception architecture (He *et al.*, 2016) demonstrated state-of-the-art performance on early benchmarks by mining mesoscopic artifacts. Similarly, Xception-based models and region-specific networks targeting eyes or lips (Haliassos *et al.*, 2021) have further improved accuracy. More recent efforts seek to improve generalization through techniques such as adversarial training, data augmentation, and incremental learning (Ojha *et al.*, 2023; Yu *et al.*, 2022) (Ojha *et al.*, 2023). However, spatial-domain detectors remain vulnerable to compression, downsampling, and post-processing, which can obscure subtle forgery traces (Abbas & Taeihagh, 2024).

## Frequency-Domain Forgery Detection:

To overcome spatial limitations, researchers have explored frequency-domain representations, where generative pipelines often leave distinctive spectral artefacts. (Durall *et al.*, 2020) employ the Discrete Fourier Transform (DFT) to expose anomalies in high-frequency bands, while others apply Wavelet or cosine transforms to mine multiscale artifacts (Qian *et al.*, 2020; Wang *et al.*, 2020). Frank *et al.* (2020b) and Khalifa *et al.* (2022) demonstrate that fixed high-pass and Gabor filters can amplify forgery signals in the spectrum, enabling simple classifiers to distinguish real from fake images.

Building on these insights, dual-branch networks, exemplified by Qian et al. (2020), are designed to learn adaptive frequency representations through explicit separation of high- and low-frequency components, thereby enhancing model robustness. In a comparative study, Masi *et al.* (2020) demonstrated that combining color-space and frequency-domain artifacts yields complementary benefits, while Luo *et al.* (2021) introduced unified frameworks that integrate multiple high-frequency descriptors to improve generalization. Further, Yang *et al.* (2023) underscore the necessity of jointly modelling both the magnitude and phase spectra, arguing that reliance on a single spectral component renders models more susceptible to noise. By designing dedicated branches and incorporating attention mechanisms for each spectral representation, such methods have achieved improved performance across benchmark datasets including FF++ (Rössler *et al.*, 2019).

Nevertheless, most existing approaches remain constrained using conventional transforms such as the DFT, Discrete Cosine Transform (DCT), or wavelets. These methods inherently lack the directional sensitivity and fine-grained geometric localization required to capture the anisotropic and curved structures prevalent in facial manipulations. The Curvelet Transform, however, provides a multiscale representation with superior directional selectivity and edge localization, attributes well aligned with the subtle and spatially distributed artifacts introduced by deepfake synthesis. While Curvelets have demonstrated success in various domains such as image denoising (Zhao *et al.*, 2024), medical imaging (Nayak *et al.*, 2017), and copy-move forgery detection using traditional machine learning with texture analysis approaches (Al-Hammadi *et al.*, 2013), their integration with deep learning methods for deepfake detection remains unexplored. This is particularly notable given the greater complexity and adversarial resilience of deepfakes compared to conventional forms of image manipulation.

To address this gap, our work investigates the utility of Curvelet-based representations in encoding discriminative features for synthetic facial content and their integration within a CNN-based detection pipeline. In doing so, we extend frequency-aware deepfake detection research by incorporating a transform inherently suited to capturing the curved, oriented structures that typify facial forgeries.

## Methodology:

This section details the components of our proposed approach. We begin with an overview of the full pipeline, followed by descriptions of the two core modules: the wedge-level attention mechanism (WedgeSE) and the multiscale enhancement strategy (Scale Module), each designed to enhance the Curvelet-based frequency representations for improved deepfake detection.

## Overview:

Figure 1 illustrates the main steps of the proposed approach: Panel (a) shows the overall pipeline; Panel (b) shows the processing of one channel. $4 \times H \times W$ denote the four reconstructed feature representations via inverse 2D Fast Discrete Curvelet Transform (iFDCT), $H$ and $W$ denote the height and width of the input image. Our method operates on each RGB channel independently. Given RGB colour channel $C$, we first transform C into 42 wedge coefficients using the 2D Fast Discrete Curvelet Transform (FDCT), each corresponding to a specific scale and orientation. We then decompose the magnitude and phase components of each wedge, and modulate the magnitude while keeping the phase as it is.

We modulate the magnitude in two ways: First, to adaptively highlight the most discriminative frequency components, we incorporate a wedge-specific attention mechanism, which consists of a hierarchical sequence of six depthwise convolutional layers. It (i) looks at each curvelet wedge, (ii) scores how useful that wedge is for detecting forgery, and (iii) turns the score into a simple on/off gate (binarized, 0 or 1) via thresholding. Since the score is computed from the wedge's own spatial map, the gate is spatially aware, letting the network favour specific directions and scales where informative wedges, those showing subtle texture or periodic artifacts, are kept and amplified; unhelpful wedges are muted. Second, we define a Scale Masks module that partitions the spectrum of wedges into four frequency bands, referred to as scales. This is done by each scale receiving an identical copy of all gated wedges coming from WedgeSE, and independently applying a combination of fixed binary masks and learnable spatial masks to its copy. After modulation, each scale will have its own resulting set of magnitude-enhanced wedges. We recompose each wedge of each scale by combining its modified magnitude with its original phase, and apply the iFDCT to reconstruct four spatial feature maps for the given colour channel. Repeating this process for each of the three RGB channels yields 12 enhanced representations in total. These are concatenated and passed to a modified, pretrained Xception network, which is fine-tuned end-to-end for binary deepfake classification. The following subsections provide a detailed description of each key component in this architecture.

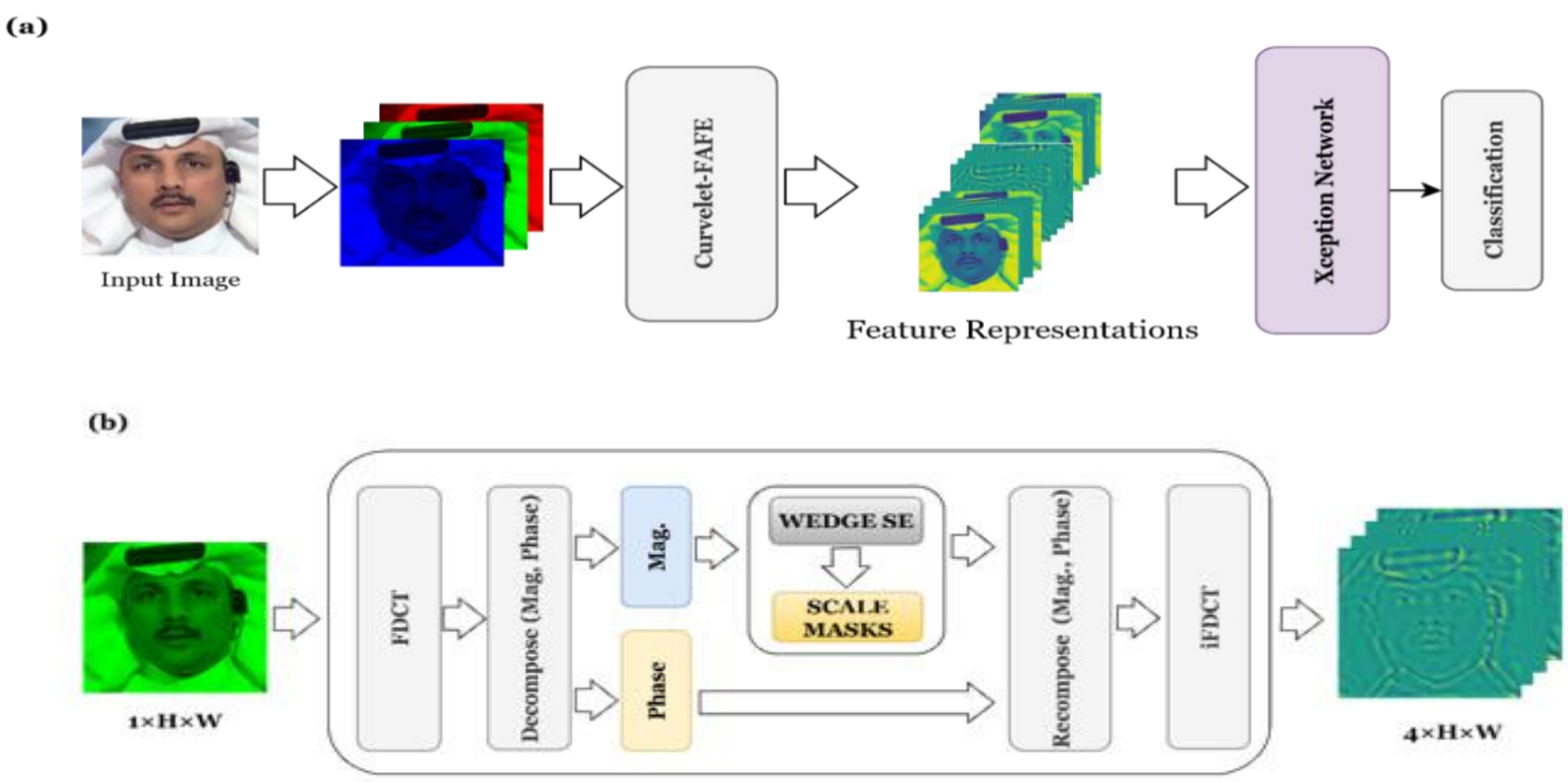

**Figure 1:** (a) High-level Overview of the proposed approach pipeline. (b) Procedure of Curvelet-FAFE on a single RGB channel.

## Curvelet Transform:

Our implementation utilizes the FDCT as introduced by Candès *et al.* (2006). The FDCT partitions the frequency domain into multiple angular components across scales, generating coefficients that encode localized directional frequency content. Coarser scales capture global, low-frequency components, while finer scales represent detailed, high-frequency features. In our configuration, we employ five scales with an initial number of angles set to eight, resulting in 42 frequency representations per RGB channel, known as wedges. The input resolution for our curvelet system is set to 299×299 pixels to match the input image size. We summarize the distribution of curvelet responses across scales as follows:

- Scale 1: The coarsest scale, containing a single wedge that captures global low-frequency information.
- Scale 2: Eight wedges capturing low-frequency content at distinct orientations.
- Scales 3 and 4: Each with 16 wedges, representing higher frequency bands with finer angular resolution.
- Scale 5: The finest scale, consisting of a single wedge encoding the highest-frequency components.

To facilitate learning, and since each curvelet coefficient is originally a complex number representing local frequency and orientation content at a specific scale and direction, we write the coefficient as:

$$z_{j,\ell,k} = a + b = mag \cdot (\cos\theta + i\sin\theta) \qquad (1)$$

Where $z_{j,\ell,k}$ represents the complex Curvelet coefficient at scale $j$, orientation (wedge) $\ell$, and spatial location $k$
$a$ and $b$ denote the real and imaginary parts of the coefficient respectively
$mag$ denotes the magnitude of the coefficient
$\theta$ is the phase angle representing the local orientation

We then decompose each complex coefficient $z_{j,\ell,k}$ into magnitude and phase components using:

$$mag = \sqrt{a^2 + b^2}, \qquad \theta = phase = \tan^{-1}(\frac{b}{a}) \qquad (2)$$

The magnitude is subsequently passed through our proposed gating and scale modules to compute a gated and recalibrated magnitude, denoted as $mag_g$. Finally, we re-compose the modified magnitude with the original phase to form new complex coefficients:

$$\hat{z}_{j,\ell,k} = mag_g \cdot e^{i\theta} = mag_g \cdot (cos\theta + isin\theta) \qquad (3)$$

Prior works in image processing and computer vision (Zhao *et al.*, 2024; Al-Hammadi *et al.*, 2013) commonly reduce the number of curvelet wedges or downsample their spatial resolution to mitigate computational demands. In contrast, we preserve the full wedge decomposition at the original spatial resolution, allowing the model to learn which frequency-orientation components are most informative down to the sample level. This design choice is motivated by the observation that deepfake artifacts often appear inconsistently across scales and directions (Rössler *et al.*, 2019), making selective or coarse analysis potentially ineffective. The next subsection introduces our proposed mechanism for learning adaptive wedge selection: WedgeSE.

## WedgeSE: Wedge-level Attention Mechanism

Squeeze-and-Excitation (SE) blocks, introduced by Hu *et al.* (2018), are widely utilized in modern convolutional architectures due to their ability to enhance feature representation by modelling inter-channel dependencies. Their success in many vision tasks can be attributed to two main presumptions (Hu *et al.*, 2018; Woo *et al.*, 2018): (1) spatial details can be compressed via global average pooling into global channel descriptors without significantly degrading information, and (2) their typical placement in intermediate layers where feature maps are semantically rich and spatially reduced. However, these presumptions become problematic in the context of curvelet-transformed frequency features. Applying average global pooling to the curvelet wedges would inevitably discard vital, spatially anchored information, given its directional and multiscale nature.

To address this, we introduce WedgeSE, a modified squeeze-and-excite gating mechanism specifically designed for curvelet wedges. Figure 2 illustrates the main steps of the WedgeSE module. First, for the Squeeze operation: instead of collapsing spatial dimensions via global pooling, WedgeSE uses a squeeze block of a hierarchical sequence of six depthwise convolutional layers, each with a stride of 2, which progressively reduce the spatial size of each wedge from 299×299 down to 4×4. The kernel sizes are set to 5 for the first two layers, and 3 for the remaining layers. Zero-padding of 1 is applied in the first four layers, while the last two layers use no padding. This carefully designed padding strategy ensures that the final convolutional layer produces an exact 4×4 feature map while minimizing the influence of artificial border values on the extracted features. A final global average pooling layer then produces a single scalar per wedge. Second, to enable non-linear modelling of wedge importance, the pooled features are passed through an excite block, which consists of a lightweight two-layer MLP with a ReLU activation in between, and a sigmoid activation at the end, to produce attention-based single scalar per wedge weights. Thresholding is applied to convert sigmoid values to 1 for values larger than or equal to 0.5, and to 0 otherwise.

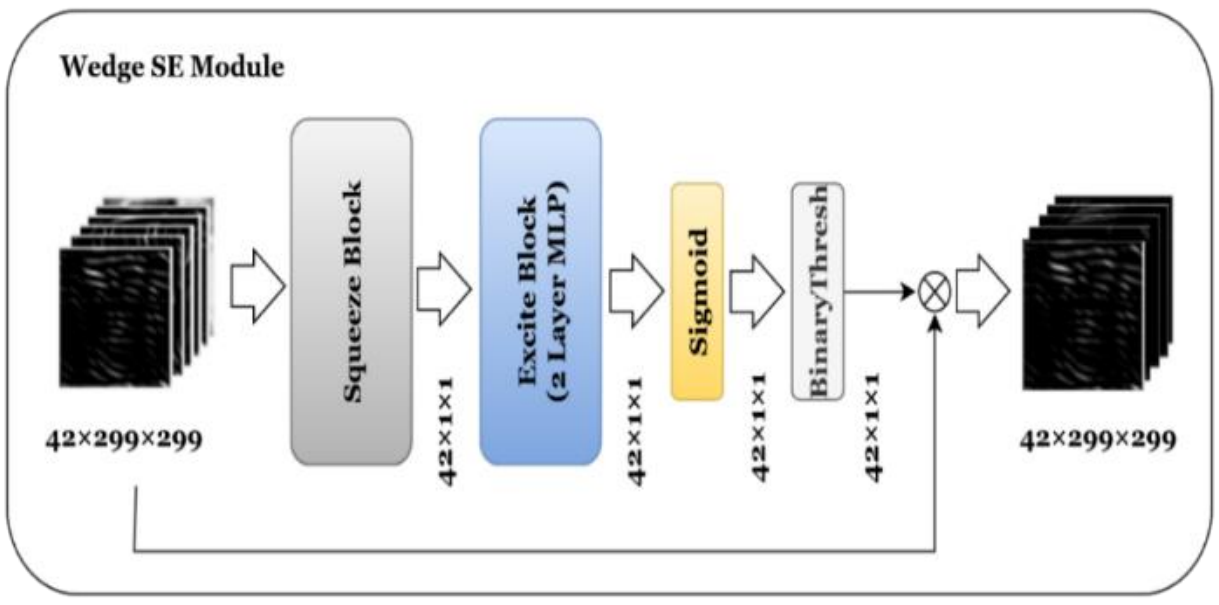

**Figure 2:** The main steps of the WedgeSE module.

Mathematically, given wedge coefficients $x \in \mathbb{R}^{C\times H\times W}$, where each wedge is treated as a separate channel, the gating vector $g \in \mathbb{R}^{C\times 1\times 1}$, is computed as:

$$g_{C\times 1\times 1} = \sigma(MLP(f_{pool}\left(f_{depthwise}(x)\right))) \quad (4)$$

Where $g_{C\times 1\times 1} \in (0,1)$ denotes the per channel importance weights
$f_{depthwise}$ denotes the stack of depthwise convolutions
$f_{pool}$ is a global average pooling operation to $1 \times 1$
$\sigma$ is the sigmoid function used to constrain the gating values between 0 and 1

Each gating value is then binarized through thresholding to zero out the magnitude of certain wedges via element-wise multiplication, effectively learning deepfake-method-specific, sample-level cues to find a global set of most informative wedges for deepfake detection:

$$\check{X}_{C\times H\times W} = binary_thresh(g_{C\times 1\times 1}) \odot X_{C\times H\times W} \quad (5)$$

Where $g_{C\times 1\times 1} \in (0,1)$ denotes the per channel importance weights
$\check{X}_{C\times H\times W}$ denotes the recalibrated wedges
$\odot$ denotes the element-wise multiplication
$binary_thresh$ denotes binary thresholding of gate weights to 1 or 0

However, during training, a recurring challenge with the WedgeSE gating mechanism is its propensity to saturate by activating a majority, if not all, wedges prematurely during training. To mitigate this, we introduce a progressively intensified L1 regularization strategy that explicitly encourages sparse gate activations. This regularization is designed to be classification loss-sensitive, adapting dynamically throughout training: it relaxes when the loss is high, allowing exploratory activation of gates, and tightens as performance stabilizes, thereby ensuring that the enforcement of sparsity does not come at the expense of predictive accuracy.

Mathematically, the L1 loss quantifies the mean absolute deviation between active gate counts per channel and a target activation proportion and is combined with a normalized classification loss term to dynamically adjust the regularization intensity in response to the model's learning progress. let $L_{cls}$ denote the classification loss, the normalized classification loss is calculated as follows:

$$\hat{L}_{cls} = min(\lambda_{max}, max\left(\lambda_{max}, \frac{L_{cls} - L_{min}}{L_{max} - L_{min}} \cdot \lambda_{max}\right)) \quad (6)$$

Where $\hat{L}_{cls}$ denotes the normalized classification loss
$\lambda_{max}$ is a scaling factor that sets the maximum allowable contribution of the normalized classification loss to the total regularization
$L_{min}$ and $L_{max}$ are empirically defined lower and upper bounds for $L_{cls}$ used to normalize its values into a range between 0 and $\lambda_{max}$.
$\lambda_{cls}$ is a weighting factor for the classification loss contributing to the classification
$\hat{L}_{cls}$ denotes the normalized classification loss
$\hat{L}_{cls} \in [0, \lambda_{max}]$ ensures the regularization impact is bounded and does not grow arbitrarily large.
To calculate L1 regularization, Let $L_1^R,\ L_1^G, the\ L$ denote the L1 norms of the gate activations for the Red, Green and Blue channels, respectively. The regularization is defined as:

$$L_{reg} = (\frac{1}{3}\sum\nolimits_{c \in \{R,G,B\}} |L_1^c - T| \cdot \lambda_{L1}) + (\lambda_{cls} \cdot \hat{L}_{cls}),\ \ T = \frac{1}{3} \cdot M \quad (7)$$

Where $T$ is the target number of active gates per channel
$M$ denotes the total number of allowed active gates across all RGB channels
$L_1^c$ denotes L1 gate activation per RGB channel c
$\lambda_{L1}$ is a scalar hyperparameter controlling the strength of L1 sparsity regularization

$\lambda_{cls}$ is a weighting factor for the classification loss contributing to the classification
$\hat{L}_{cls}$ denotes the normalized classification loss as defined in equation (8)
$L_{reg}$ is the total regularization loss combining sparsity and performance sensitivity
In our implementation, we set $L_{min} = 0.2$, $L_{max} = 0.5$, and $\lambda_{max} = 0.25$, $\lambda_{L1}$ and $\lambda_{cls}$ set empirically to 0.01, and 0.1 respectively.
This mechanism ensures the model avoids over-reliance on one channel and learns compact, complementary representations. The gated wedges are then passed on to our proposed Scale module which is explained in the next subsection.

## Scale Masks: Multiscale Enhancement Strategy:

Inspired by the Frequency-Aware Decomposition module (FAD) in Qian *et al.* (2020), we design a Scale module that divides the 42 curvelet wedges into contiguous frequency bands, which we refer to as scales. To simplify processing, we remap the original five curvelet scales into three new frequency scales, where each scale represents a frequency band, from Coarse to High, with a fourth global scale to represent all frequency bands.

Each scale receives an identical copy of the 42 curvelet wedges and maintains two types of masks for every wedge: fixed base masks and learnable masks, both matching the wedges in resolution and number. For every base mask, if its corresponding wedge falls in the range of the frequency band of that scale, then it is assigned a value of 1, and 0 otherwise. For example, in Scale 1, for wedges ∈{1, 2, …, 9}, the corresponding base masks are set to 1s, while the remaining 33 wedges are set to 0s. In Parallel, learnable masks are enabled for all wedges. These learnable masks are first passed through a scaled sigmoid function, which bounds the learnable values between -1 and 1 to allow both attenuation and amplification of curvelet wedge magnitudes. After that, the learnable masks are added element-wise to the base masks to form the final modulation masks. Table 1 presents the new scale mapping of wedges, base masks and learnable masks, and their distribution for each scale.

**Table 1:** The Mapping of Curvelet Scale Groups to our proposed Redefined Scale Groups.

| Scale (band) | Curvelet Scales Represented | 1 Valued Base Masks | 0 Valued Base Masks | Learnable Masks |
|---|---|---|---|---|
| Scale 1 (Coarse) | 1 & 2 | 9 | 33 | 42 |
| Scale 2 (Low–Mid) | 3 | 16 | 26 | 42 |
| Scale 3 (Mid–High) | 4 & 5 | 17 | 25 | 42 |
| Scale 4 (All Frequency Bands) | all | 42 | 0 | 42 |

Once the final masks are computed, each mask is then multiplied element-wise with its corresponding wedge. Figure 3 presents the process of combining and multiplying the masks with the curvelet wedges for a single scale.

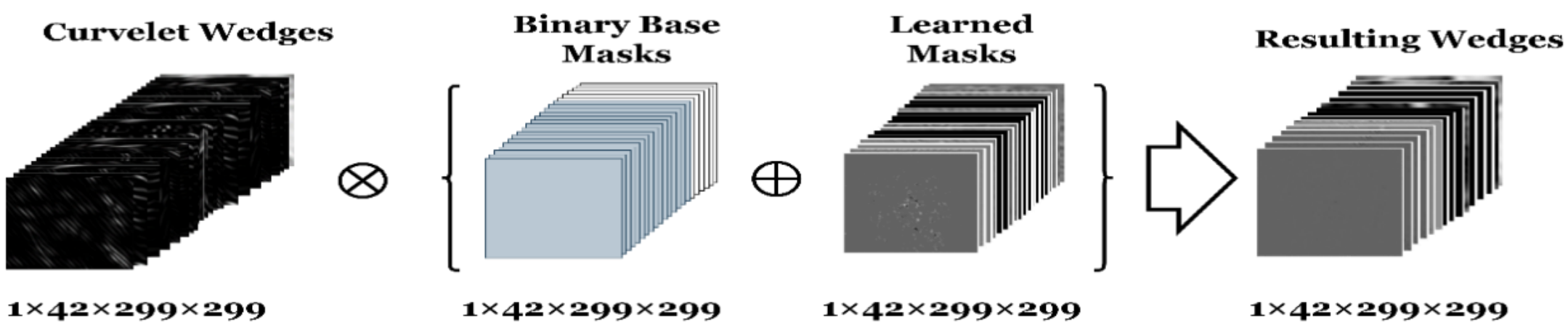

**Figure 3:** The process of combining and multiplying the base and learnable masks with curvelet wedges for a single scale.

Mathematically, let the 42 wedges be denoted as:

$$U = \{\omega_i \in \mathbb{R}^{H\times W} \mid i = 1, \dots, 42\} \quad (8)$$

Where $U$ is the set of all 42 curvelet wedges for a given RGB channel.
$\omega_i$ denotes the *i*-th wedge of curvelet coefficients of a specific scale and orientation
$\mathbb{R}^{H\times W}$ denotes the space of real-valued matrices with wedge spatial dimensions height $H$ and width $W$
$i$ is the index corresponding to each wedge

We define the new frequency scales, each denoted as:

$$S_k \subseteq \{1, \dots, 42\}, \quad \text{for } k = 1, \dots, 4 \quad (9)$$

Where $S_k$ denotes the set of wedge indices that belong to frequency scale (band) $k$.
$k$ denotes the index of the four predefined frequency bands in the Scale module

For each scale $S_k$, we define:

1. Base masks:

$$B_k \in \{B_k[i] \in \mathbb{R}^{H\times W} \mid i = 1, \dots, 42\}:$$

$$B_k[i] = \begin{cases} 1_{H\times W} & if\ i \in S_k \\ 0_{H\times W} & otherwise \end{cases} \quad (10)$$

where $B_k$ is the set of base masks for scale k, one per wedge
$H \times W$ are the spatial dimensions of the wedge.

$B_k[i] \in \mathbb{R}^{H\times W}$ is the binary mask corresponding to wedge $i$ in scale $k$

$1_{H\times W}$ and $0_{H\times W}$ are all-ones and all-zeros matrices, respectively

2. Learnable spatial masks:

$$M_k \in \{M_k[i] \in \mathbb{R}^{H\times W} \mid i = 1, \dots, 42\} : \quad (11)$$

where $M_k$ is the set of modulation masks for scale k, one per wedge

$M_k[i] \in \mathbb{R}^{H\times W}$ is the learnable modulation mask for wedge $i$ in scale $k$

3. Final mask $C_{k,i} \in \mathbb{R}^{H\times W}$, we define:

$$C_{k,i} = \sigma(M_k[i]) + B_k[i] \quad (12)$$

where $C_{k,i} \in \mathbb{R}^{H\times W}$ is the final modulation mask for wedge $i$ in scale $k$

$\sigma$ is the scaled sigmoid activation, defined as $\sigma(x) = 2 \cdot sigmoid(x) - 1$

The modulated magnitude of wedge $i$ in scale $k$ is then:

$$\breve{U}_{k,i} = C_{k,i} \odot U_{k,i} \quad (13)$$

where $U_{k,i}$ is the $i$-th wedge in scale $k$ as defined in Equation (8)

$\breve{U}_{k,i}$ is the altered wedge $i$ in scale $k$

$\odot$ denotes element-wise multiplication.

Finally, for each scale $S_k$, we recompose the magnitude and phase of the filtered, using Equation (3) and as follows:

$$\breve{S}_k = \{ \breve{U}_{k,i} \odot e^{j\theta_i} \mid i \in S_k \} \quad (14)$$

where $\breve{S}_k$ is the scale $k$ containing recomposed wedges

$e^{j\theta_i}$ denotes the phase of wedge $i$,

$j$ is the imaginary unit, satisfying $i^2 = -1$

The inverse fast discrete curvelet transform (iFDCT) is then applied to $\breve{S}_k$ resulting in 4 single channel 2D feature representations for each RGB channel. Figure 4 shows the four single channel feature representations for a single RGB channel.

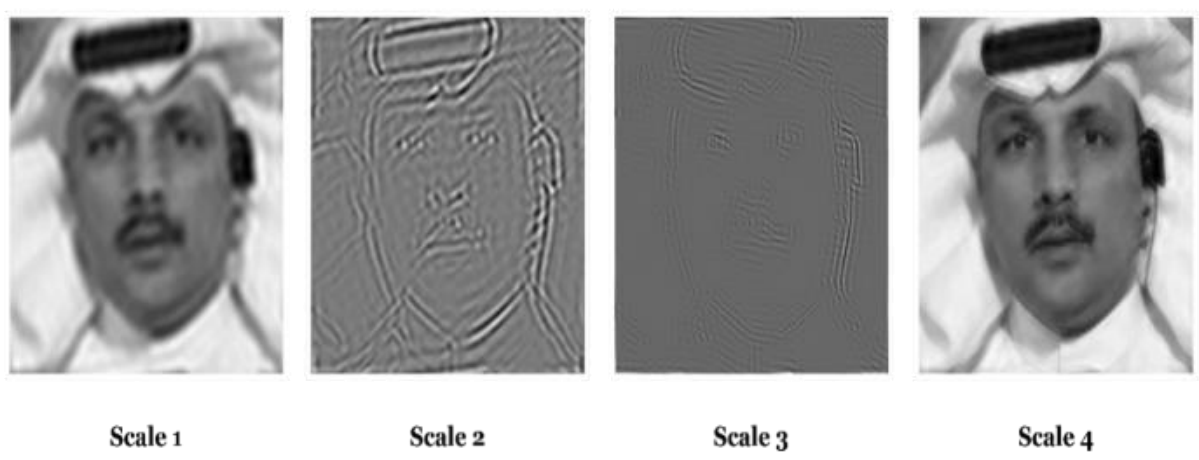

**Figure 4:** Result of Curvelet-FAFE for a single RGB color channel.

The newly formed 12-channel feature representations are concatenated and passed to a modified, pretrained Xception network, which is fine-tuned end-to-end for binary deepfake classification.

## Experiments:

In this section, we introduce the overall experimental setups. Then, we present a comprehensive evaluation of the proposed approach covering various aspects, including datasets, implementation details, performance, and other details to be described.

## Dataset:

Following prior face forgery detection works, we conduct our experiments on the FaceForensics++ (FF++) dataset (Rössler *et al.*, 2019), a widely used benchmark for deepfake detection. FF++ contains 1,000 real videos, collected from YouTube with subject consent, primarily featuring frontal, unobstructed faces. The dataset is split into 720 training videos, 140 for validation, and 140 for testing.

Each real video is manipulated using four distinct forgery methods. **DeepFakes (DF)** uses an autoencoder-based pipeline to replace a source face with that of a target identity while maintaining the original expressions (DeepFakes GitHub, 2018). **Face2Face (F2F)** transfers expressions from one person to another in real time through 3D morphable models (Thies *et al.*, 2016). **FaceSwap (FSwap)** takes a more graphics-based approach, aligning, warping, and blending a source face into the target frame (FaceSwap GitHub, 2016). Finally, **NeuralTextures (NT)** uses neural rendering to model view-dependent textures, enabling highly detailed and photorealistic facial reenactment (Thies *et al.*, 2019). In total, FF++ offers 5,000 videos, each containing between 300 and 700 frames, covering a broad spectrum of manipulations that make it an ideal testbed for developing and evaluating detection methods.

## Implementation Details:

Dlib (King, 2009) is employed to detect faces in each frame. The largest bounding box is selected and expanded by 30% on each side, following the procedure described in Haliassos et al. (2021). The resulting region is resized to 299×299 pixels to match the input resolution of our backbone. From each video, we extract the first 300 frames and divide them into four equal-length segments. During training, one frame is randomly sampled from each training segment, while the first frame in each segment is sampled during validation and testing. To mitigate class imbalance, real videos are over-segmented fourfold during training, but equally segmented during validation and testing.

The backbone of our model is the Xception network (Chollet, 2017), pretrained on ImageNet-1K. To accommodate our 12-channel input representation, we adapt the first convolutional layer by extending the original 3-channel RGB weights. Specifically, we duplicate the pretrained weights corresponding to the R, G, and B channels, four times each, to initialize the 12-channel input filter, preserving the semantic structure of the original initialization. Additionally, the final classification layer is replaced with a binary output layer suitable for deepfake detection The model is optimized using Adam (Kingma & Ba, 2015) with a learning rate of 0.002, cosine learning rate decay, a weight decay of $1e-4$, and a batch size of 8. L1 Regularization initially set to $2.5e-4$ and increases by half of the initial value every 5 epochs.

For evaluation, we adopt Accuracy (Acc) and the Area Under the Receiver Operating Characteristic Curve (AUC) as our primary metrics. Following Rössler et al. (2019), accuracy is measured by averaging the frame-level scores across each video, and is defined as:

$$Acc = \frac{TP + TN}{TP + TN + FP + FN} \quad (15)$$

where $TP$, $TN$, $FP$, $FN$ denote true positives, true negatives, false positives, and false negatives, respectively.

AUC is computed in the same manner, as done in Face X-ray (Li *et al.*, 2020), to better reflect model confidence and robustness in distinguishing real from fake content.

$$AUC = \int_0^1 TPR(FPR^{-1}(x))\,dx \tag{16}$$

where $TPR$ denotes true positive rate, and $FPR$ denotes false positive rate, respectively.

## Evaluation:

In this section, the performance of our proposed method in comparison with existing deepfake detection techniques is evaluated. Hence, this is achieved through two official compression levels of the FF++ dataset: High Quality (HQ) represents low compression, and Low Quality (LQ) represents high compression.

Table 2 summarizes the accuracy and AUC metrics compared against several state-of-the-art methods. On the HQ subset, our proposed Curvelet-FAFE approach achieves an accuracy of 98.48% and an AUC of 99.96%, demonstrating superior discriminative performance and surpassing all existing methods in accuracy and AUC. For the LQ subset, which is more challenging due to heavy compression artifacts, our method attains 89.93% accuracy and 92.75% AUC, placing it competitively within the upper tier of current approaches. Notably, while F3-Net (Qian *et al.*, 2020) achieves the highest accuracy and AUC on LQ data, primarily due to its Local Frequency Statistics module (LFS), and Mix-Block attention module, our method maintains robust performance, highlighting its robustness to compression-induced distortions despite relying solely on lightweight attention and curvelet frequency cues.

**Table 2:** Comparison of detection performance (Accuracy [%] and AUC [%]) on the FF++ dataset under two compression settings: HQ (low compression) and LQ (high compression). Bold and underlined results indicate first and second best, respectively.

| Method | FF++ (HQ) | | FF++ (LQ) | |
|---|---|---|---|---|
| | Acc. | AUC | Acc. | AUC |
| Xception (Chollet, 2017) | 95.73 | 96.30 | 86.86 | 89.30 |
| Xception-ELA (Gunawan *et al.*, 2017) | 93.8 | 98.40 | 79.63 | 82.90 |
| Xception-PAFilters (Chen *et al.*, 2017) | - | - | 87.16 | 93.30 |
| Face X-ray (Li *et al.*, 2020) | - | 87.35 | - | 61.60 |
| F3-Net (Qian *et al.*, 2020) | <u>97.52</u> | 98.10 | **90.43** | **93.30** |
| FDFL (Li *et al.*, 2021) | 96.69 | <u>99.30</u> | 89.00 | 92.40 |
| **Curvelet-FAFE (ours)** | **98.48** | **99.96** | <u>89.93</u> | <u>92.75</u> |

## Grad-CAM Visualization:

To further understand the discriminative behaviour of our proposed approach, Grad-CAM visualization is employed (Selvaraju et al., 2016). The saliency maps are shown in Figure 5. Across most deepfake methods, the model attends to multiple regions of the face, reflecting the widespread presence of manipulation artifacts. However, In the case of NeuralTextures forgeries, the saliency maps consistently highlight the lips and surrounding lower facial areas. This is expected, as NeuralTextures performs expression transfer that is limited to the mouth region, using UV-based reenactment techniques, without altering other facial areas (Dang & Nguyen, 2023). This pattern suggests that Curvelet-FAFE effectively captures localized frequency inconsistencies, adapting its focus to the characteristic weaknesses of each manipulation method.

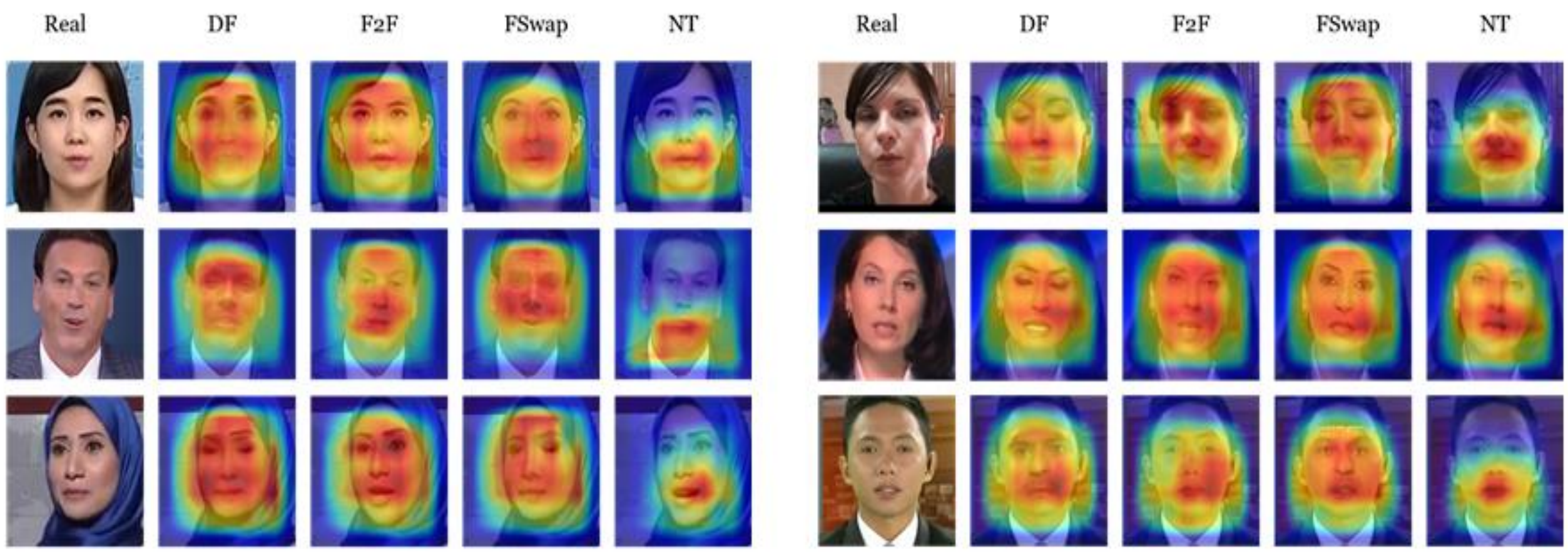

**Figure 5:** Saliency maps for different deepfake methods using Grad-CAM (Selvaraju *et al.*, 2016) visualization.

**Comparison of Training and Validation Performance:**

Figure 6 shows the training and validation curves for loss and accuracy. To further analyse the learning behaviour of the model, the training and validation curves for loss and accuracy on the FF++ (HQ) Dataset are presented. The training and validation curves reflect a consistent and well-structured learning trajectory. Training accuracy improves while validation accuracy follows closely, indicating strong generalization. Training loss steadily declines, with validation loss mirroring this trend. Minor fluctuations in mid-epochs are observed, likely reflecting adaptation to complex features. Overall, the results demonstrate robust convergence, effective feature learning, and strong generalization across epochs.

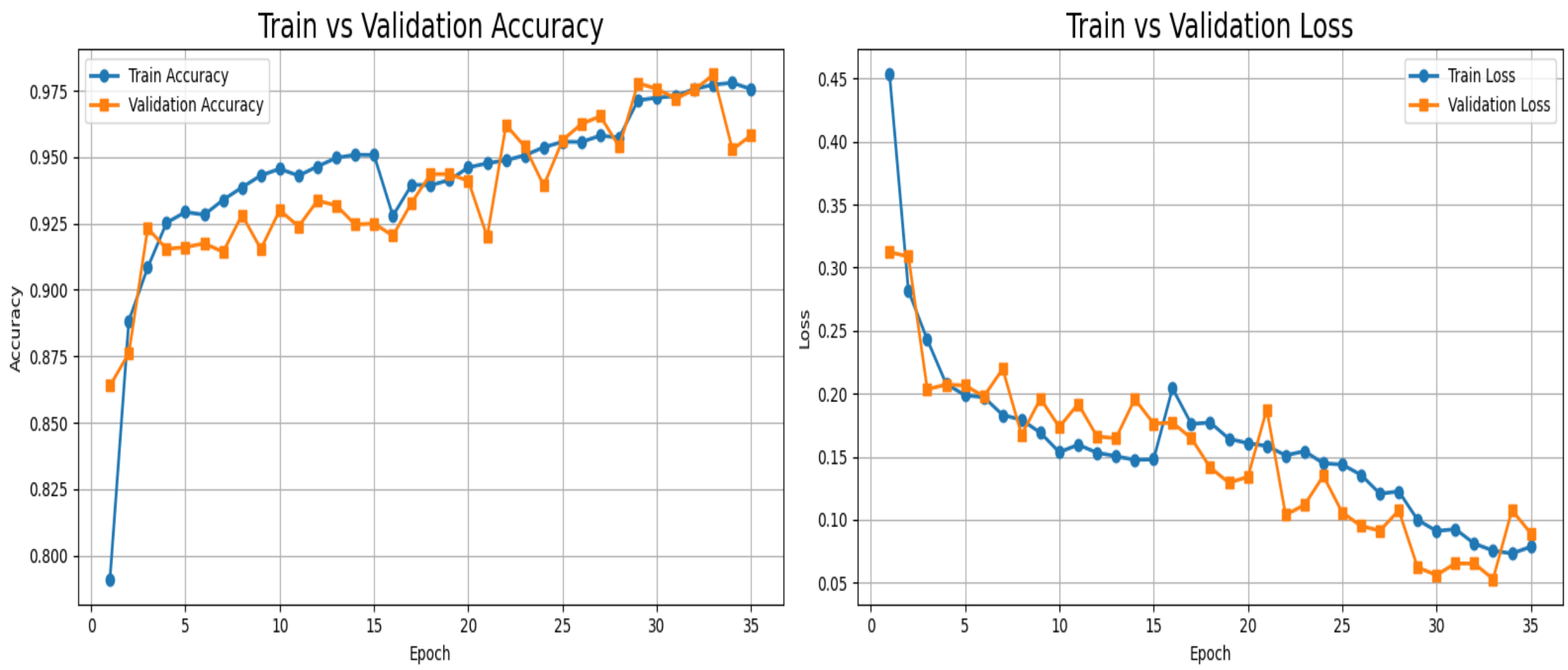

**Figure 6:** Train vs Validation accuracy (Left) and loss (Right) curves.

## CONCLUSION

In this work, we proposed Curvelet-FAFE, a frequency-aware deepfake detection approach that exploits the multiscale and directional properties of the Curvelet Transform. By decoupling magnitude and phase, and introducing wedge-level gating with spatial awareness, the model adaptively highlights discriminative frequency components. A customized, progressively intensified, loss-sensitive L1 regularization promotes sparsity and balanced attention across RGB channels without sacrificing performance. Experiments on the FF++ dataset demonstrate that Curvelet-FAFE performs robustly across varying levels of compression, achieving competitive AUC and learning artifact patterns. Nonetheless, this work has several limitations. The current design primarily leverages magnitude information and treats RGB channels independently. However, incorporating phase components, which are essential for capturing subtle temporal and structural inconsistencies, may further improve detection performance. Second, while the method incorporates all RGB channels, their joint interactions could be explored more deeply to enhance detection further. Third, our evaluation is confined to FF++, and cross-dataset generalization remains to be validated, which is crucial for real-world deployment. In future work, we plan to address these limitations by prioritizing the most informative wedge patterns, refining cross-channel integration, and enhancing efficiency through evaluation across multiple datasets and integration with alternative backbones and multi-modal input.

**Acknowledgement:**

The authors gratefully acknowledge the support provided by the University of Zakho, Kurdistan Region, Iraq.

**Ethical Statement:**

This study has used existing public breast cancer datasets. No experiments have been done on humans.

**Author Contributions:**

S.A.S. conceptualized the study, designed the methodology, data analysis, and wrote the original draft. R.J.M. verified the results, edited the manuscript, and performed a review of the manuscript. Both authors read and approved the final version of the manuscript.

**Conflict of Interest:**

The authors declare that there are no conflicts of interest regarding the publication of this paper.

**Funding:**

None.